\documentclass[10pt,twocolumn,letterpaper]{article}

\usepackage{cvpr}              % To produce the CAMERA-READY version
\usepackage[accsupp]{axessibility}
\usepackage{graphicx}
\graphicspath{ {./media/} }
\usepackage{amsmath}
\usepackage{amssymb}
\usepackage{booktabs}

\usepackage{wrapfig}
\usepackage{enumerate}
\usepackage{enumitem}
\usepackage{tabularx}
\usepackage{longtable}
\usepackage{ltablex}
\usepackage{makecell}

\usepackage[pagebackref,breaklinks,colorlinks,citecolor=cvprblue]{hyperref}
\usepackage{multirow}

\usepackage{xcolor}

\newcommand{\ours}{\textsc{LaMenDa}\xspace}

\newcommand{\pixtostruct}{Pix2Struct\xspace}
\newcommand{\matcha}{\textsc{MatCha}\xspace}
\newcommand{\deplot}{\textsc{DePlot}\xspace}

\usepackage[capitalize]{cleveref}
\crefname{section}{Sec.}{Secs.}
\Crefname{section}{Section}{Sections}
\Crefname{table}{Table}{Tables}
\crefname{table}{Tab.}{Tabs.}
\Crefname{figure}{Figure}{Figures}
\crefname{figure}{Fig.}{Figs.}

\usepackage{xspace}
\makeatletter
\DeclareRobustCommand\onedot{\futurelet\@let@token\@onedot}
\def\@onedot{\ifx\@let@token.\else.\null\fi\xspace}

\def\eg{\emph{e.g}\onedot} 
\def\ie{\emph{i.e}\onedot}

\makeatother

\definecolor{cvprblue}{rgb}{0.21,0.49,0.74}

\def\paperID{5769} % *** Enter the Paper ID here
\def\confName{CVPR}
\def\confYear{2024}

\begin{document}

% \title{Synthetic Data Improves Visual Reasoning on Chart Plots}
\title{Synthesize Step-by-Step: Tools, Templates and\\ LLMs as Data Generators for Reasoning-Based Chart VQA}

% \and

% \author{%
% Zhuowan Li\textsuperscript{$1*$} \qquad Bhavan Jasani \textsuperscript{$2**$}  \qquad Peng Tang \textsuperscript{$2$} \\ 
% \enspace Shabnam Ghadar\textsuperscript{$2$} \\ 
% % \vspace{0.3em}
% {\textsuperscript{$1$} Johns Hopkins University    } \qquad
% {\textsuperscript{$2$} AWS AI labs} \qquad \\ \\
% % \vspace{-.25em} 
% % {\tt \small zli110@jhu.edu \enspace \{zli110}@amazon.com}
% }
\author{Zhuowan Li$^{1}$ \thanks{indicates equal contributions. The work was done when Zhuowan Li was an intern at Amazon.}  \quad Bhavan Jasani$^{2}$ \footnotemark[1] \quad Peng Tang$^{2}$ \quad Shabnam Ghadar$^{2}$ \\
$^{1}$ Johns Hopkins University \quad
$^{2}$ AWS AI Labs\\
{\tt\small zli110@jhu.edu \{bjasani, tangpen, shabnam\}@amazon.com}
}

\maketitle
%%%%%%%%% MAIN BODY
\begin{abstract}

Understanding data visualizations like charts and plots requires reasoning about both visual elements and numerics. Although strong in extractive questions, current chart visual question answering (chart VQA) models suffer on complex reasoning questions. 
In this work, we address the lack of reasoning ability by data augmentation. We leverage Large Language Models (LLMs), which have shown to have strong reasoning ability, as an automatic data annotator that generates question-answer annotations for chart images.
The key innovation in our method lies in the \textit{Synthesize Step-by-Step} strategy: our LLM-based data generator learns to decompose the complex question into step-by-step sub-questions (rationales), which are then used to derive the final answer using external tools, \ie Python. This step-wise generation procedure is trained on synthetic data generated using a template-based QA generation pipeline.
Experimental results highlight the significance of the proposed step-by-step generation. By training with the LLM-augmented data (\ours), we significantly enhance the chart VQA models, achieving the state-of-the-art accuracy on the ChartQA and PlotQA datasets. In particular, our approach improves the accuracy of the previous state-of-the-art approach from 38\% to 54\% on the human-written questions in the ChartQA dataset, which needs strong reasoning. We hope our work underscores the potential of synthetic data and encourages further exploration of data augmentation using LLMs for reasoning-heavy tasks. 

\end{abstract}

\section{Introduction}

\begin{figure}[t!]
\begin{center}
    \vspace{-1.0em}
    % \fbox{\rule{0pt}{2in} \rule{.9\linewidth}{0pt}}
    \includegraphics[width=1.0 \linewidth]{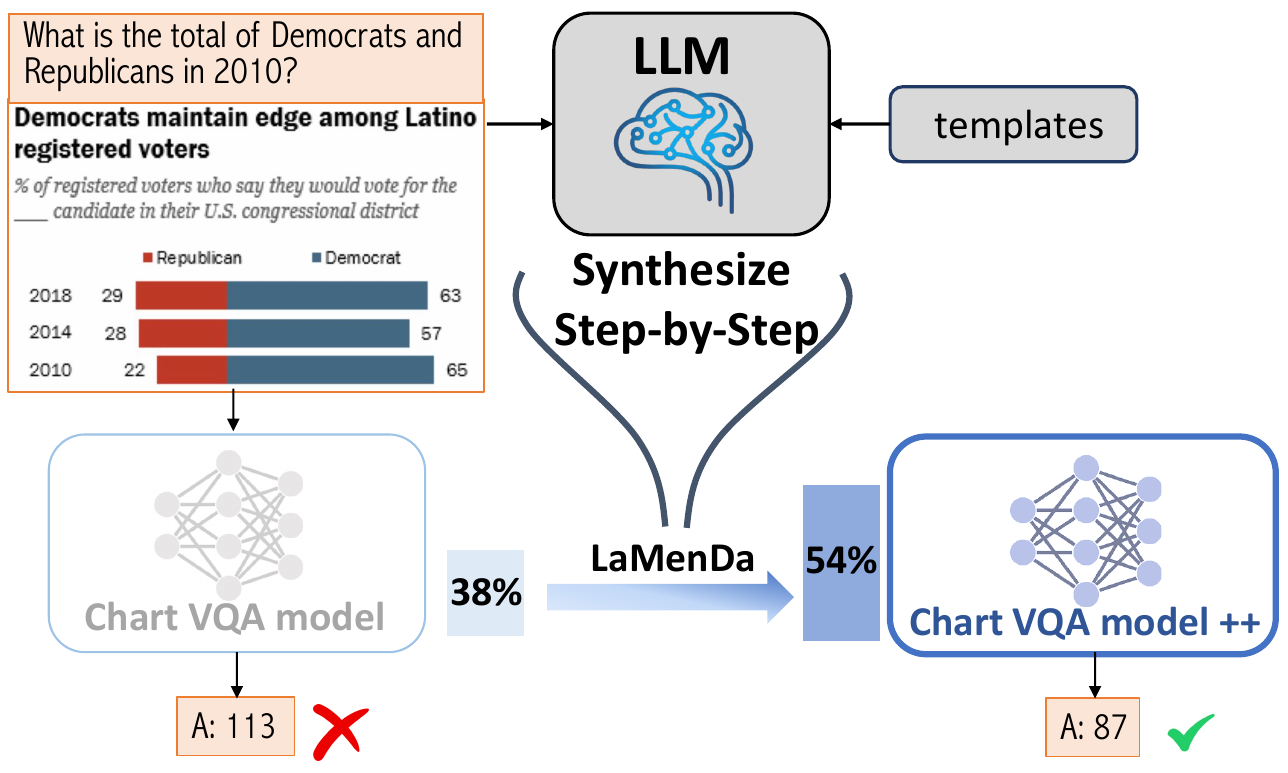}
    \vspace{-1.0em}
    \caption{Existing chart VQA models struggle with complex reasoning questions. We attribute this to limited reasoning questions in existing datasets and address it by data augmentation. We fine-tune an LLM-based data generator that automatically generates question-answer annotations given a chart image. Our key innovation is Synthesize Step-by-Step, which breaks complex questions down into easy steps that could be solved using external tools. We use templates to train the LLM. Training with LLM-augmented data, \ours, greatly enhances the chart VQA models.}
    \label{fig:introduction}
    \vspace{-2.0em}
\end{center}
\end{figure}

% \draft{para-1: chart reasoning is important, yet challenging.}
Data visualizations like charts and plots play an important role in real-world data analysis applications. Unlike natural images, chart are text-heavy and data-driven, thus requiring better visual perception (\eg OCR) and cognitive reasoning (\eg math calculation, matching the legends with bars). For example, to answer ``what is the total of Democrats and Republicans in 2010'' in \cref{fig:introduction}, the model needs to recognize the texts, match the legends of ``democrats'' and ``republicans'' to the bars, then calculate the sum. This is a challenging task requiring multimodal perception and reasoning. 

% \bhavan{This complexity also makes it difficult to ask human annotators to come up with such complex visual reasoning questions. Hence a lot of chart understanding datasets are created synthetically using some heuristics which make it easy for models to overfit and achieve high performance. While some datasets like ChartQA have human annotated question-answers they are very tiny, on an average of just 3 QAs per image, while so many QAs could be annotated. Example, the image in \cref{fig:introduction} taken from ChartQA contains only two human annotated questions - [Q: "Which color represents Republican?" A: "Red", Q: "What is the total of Republicans and Democrats in 2010?" A: "87"], even though so many questions can be asked.}

% By synthesizing step-by-step, LLM-augmented questions and answers can greatly enhance the reasoning ability of chart VQA models, improving the accuracy from 38\% to 54\%.

% Existing Chart VQA models  don't perform well on questions which involve complex reasoning about charts like the one show here. We attribute this to lack of such complex questions in the existing datasets. We propose an approach to fine-tune LLM with a vision encoder to generate question-answer pairs given an image containing chart. Just like teaching human annotator, we show it's crucial to teach LLM to generate complex questions by breaking them down into simpler sequential steps. We came up with a novel approach using templates to train the LLM. With our LLM generated data which we call LaMenDa, we demonstrate the top performing model on the challenging ChartQA dataset to improve from 38\% to 54\%.

%https://filelist.yaws.link/table_efs/users/zhuowan/ckpts/chartqa_preds/cvpr_before_after/val_human/before_after_01/#4429_303.jpeg

% \draft{para-2: describe diagnose results to say (a) current models struggle on reasoning; and (b) shortage of data.}
Compounding the complexity, the inherent difficulty of chart reasoning also amplifies the cost of collecting human annotations. Most existing datasets are synthetically generated using simple rules and heuristics, posing a risk of model overfitting. Among the datasets, only ChartQA \cite{masry2022chartqa} contains complex human-annotated question-answer pairs, although there are very few of those per image. 
For example, for the image in \cref{fig:introduction}, while a lot more questions can be asked, only two questions are annotated (What is the total of Republicans and Democrats in 2010? Which color represents Republican?). Importantly, these human-written questions pose great challenges for contemporary models. For example, the state-of-the-art model \cite{liu2022matcha} achieves only 38\% accuracy when confronted with human-written questions that require multi-step complex reasoning. Approximately half of the errors, as reported in \cite{liu2022matcha}, are due to reasoning failures.
% \footnote{The results are based on the ChartQA dataset \cite{masry2022chartqa}}. 

% \bhavan{Recently, LLMs demonstrate strong reasoning ability for solve complex tasks. We want to explore can we teach an LLM or a VLM with the instructions the way we teach human annotators? We also found it's crucial to teach the LLM/VLM to do a complex task broken down into a program of simple task which is how one usualy trains human human annotators.}

% \bhavan{We should mention that prior machine-augmented/template QAs are very easy, hence 90\%. And most of the ChartQA datasets contain mostly such QAs and have very limited human generated complex QAs.}

% \zhuowan{Not sure if this analysis is compatible with the whole story or not - keep or remove to better focus on LLM generation?} \bhavan{We should also emphasis that human annotated chart understanding datasets only have like 2-3 human annotated QAs, and so the lack of data is an issue. So we want to see if we can teach LLMs to act as trained human annotators}

% \draft{para-3: what we do - use LLMs to synthesize data step-by-step.}

In this work, we propose to leverage large language models (LLMs) as automatic data annotators for chart reasoning. Recently, LLMs have demonstrated strong reasoning ability for complex tasks \cite{wei2022chain, wang2022self, zhang2023multimodal, zhang2022automatic, yao2023tree, kojima2022large}. 
Our motivation is to take advantage of this strong reasoning ability and generate question-answer pairs using LLMs. This LLM-generated data can be used to train the chart VQA models, thus improving their reasoning ability. 
% The idea is inspired by recent works \cite{west2021symbolic, schick2021generating} that distill knowledge from LLMs into smaller downstream models.

% Given a chart image, our goal is to build a (vision-and-)language model that generate synthetic question-answer pairs for the image, which can then be used to train the downstream Chart VQA mdoels. We start with an intuitive method: we input chart image features into pretrained LLM using a projection layer (following the LLaVA \cite{liu2023visual} architecture) and tune the model to generate QA pairs. However, applying this method in this straight-forward manner yield noisy data containing incorrect answers, due to the difficulty to answer the complex reasoning questions. Alternatively, we find it critical to generate the synthetic data in a \textbf{step-by-step} manner. Instead of have the LLM generate the answers directly, we finetune it to generate the step-by-step decompositions of the question. For example, after generating the question ``what is the average between the blue and green bar?'', \todo{show this example in Fig-2 (the method figure), and mention it here.}

% For the model architecture design, we train a projection layer to project the chart image features into the LLM input space following LLaVA \cite{liu2023visual}. 

% \draft{para-4: A more detailed description of our method: In order to ..., template-based method. Then train an LLM to }
The key innovation in our method lies in the \textit{Synthesize Step-by-Step} strategy, which leverages LLMs to generate data in a step-by-step manner. Similar to human annotators who are trained to break down complex tasks into easier subtasks, our LLM-based data generator generates step-by-step. Instead of generating question-answer all at once, the generator learns to generate the question first, then decompose the complex question into step-by-step sub-questions (rationales) and answer them one by one. Finally, the answers of the sub-questions are combined to derive the final answer using external tools like calculators or Python code. Specifically, we feed the image features extracted with a ViT \cite{dosovitskiy2021image} backbone into an LLM using a simple trainable projection layer. To supervise the training of this projection layer, we construct a template-based training corpus generated using a template-based QA generation pipeline. Importantly, as we will show in the paper, the step-by-step synthesizing, trained using these template-based rationales, is critical for generating good-quality data.

% For the supervised finetuning of this step-by-step generation procedure, we first design a template-based method to generate templatic questions with step-wise decomposition raionales. Then we tune the LLM-based data generator using the templatic quesitons with step-wise rationales, and derive the answer by executing the rationales with an external calculator.

% \draft{para-5: experiment results}
Experiments showcase the effectiveness of Synthesize Step-by-Step. Our LLM-augmented Data, LaMenDa, is used to train the downstream chart VQA models and its effectiveness is shown on two challenging datasets, ChartQA and PlotQA with the state-of-the-art accuracy. Notably, we significantly improve the performance from 38\% to 54\% on the challenging human-written questions in ChartQA dataset, which requires complex reasoning. We also demonstrate the advantage of step-wise generation over straight-forward generation. With the results, we hope our work underscores the potential of synthetic data and encourages further exploration of data augmentation using LLMs for reasoning-heavy tasks.

% To summarize, our main contributions are as follows: 
% \begin{itemize}[leftmargin=*, topsep=0pt, noitemsep]
%     \item As a proof-of-concept, we augment the chart images with template-generated QA data and show that synthetic data can greatly enhance the chart reasoning on real-world data. 
%     \item We propose Synthesize Step-by-Step, which leverages LLMs as data generators that automatically generate diverse QA data with the step-wise underlying rations. We highlight that step-by-step synthesizing produces data of much better quality than the straight-forward way.
%     \item With the generated data, we show improvements over two datasets \todo{depending on whether RealCQA can work}, boosting the accuracy of existing models by more than \todo{XX\%} on the ChartQA dataset.
% \end{itemize}
% we build a (vision-and-)language model that can generate synthetic question-answer pairs given an image of a chart.

% \input{content/01_intro_bhavan}
\section{Related Work}

\noindent
\textbf{Multimodal models for text-heavy images} 
fall into two categories: OCR-dependent or OCR-free. Models like LayoutLM \cite{xu2020layoutlm}, PaLI(-X) \cite{chen2022pali, chen2023pali}, ChartBERT \cite{akhtar2023reading}, DocFormerv2 \cite{appalaraju2023docformerv2}, MQMA \cite{tang2023multiple}, DEED \cite{tang2023deed} rely on external OCR systems like ChartOCR \cite{luo2021chartocr} as either model input or training objectives. 
Recent top performing models, like Donut \cite{kim2022ocr}, Dessurt \cite{davis2022end}, Pix2struct \cite{lee2023pix2struct}, shift away from OCR thanks to vision transformers \cite{dosovitskiy2020image}. Pix2Struct \cite{lee2023pix2struct} is a transformer encoder-decoder model for multiple VQA tasks, pretrained with an HTML-masked prediction objective, using a large amount of web page screenshots. \matcha \cite{liu2022matcha} and \deplot \cite{liu2022deplot} are two recent follow-up works, tuning Pix2struct with chart-specific tasks like chart de-rendering and math reasoning, which are used in our work.
% \deplot is a chart-to-table model that translates chart images into tables. \matcha is the SoTA model for chart reasoning, which is our main evaluated model considering its top performance and modern design. 

\noindent
\textbf{Datasets for chart VQA.} 
% InfographicsVQA \cite{mathew2021infographicvqa} deals with images containing a lot of text and some figures, it doesn't necessarily focus on charts and plots. The dataset is text heavy and most of it could be solved from just text without using images. 
Earlier datasets like DVQA \cite{kafle2018dvqa}, FigureQA \cite{kahou2018figureqa}, LeafQA \cite{chaudhry2020leaf} are based on synthetically generated images and templatic questions, which are easy to solve and prior models achieve $>95\%$ accuracy. PlotQA \cite{methani2020plotqa} is also a synthetic dataset but contains open-vocabulary questions. ChartQA \cite{masry2022chartqa} is a recent dataset containing real images with human-written questions, which is much more challenging. Our work is mainly based on ChartQA and PlotQA, following the settings in \cite{liu2022matcha, liu2022deplot}.
% Following settings in MatCha \cite{liu2022matcha} primarily report performance on these datasets. One of the crucial things is the human annotated datasets are tiny even though a lot of question-answers can be created. On the other-hand the template question-answers are a lot but they lack diversity and hence model could easily overfit on them.

\noindent
\textbf{LLMs for multimodal tasks.}
There is a surge of interest in extending LLMs into the vision-and-language domain. Recent works align the image features into the input space of the pretrained LLMs using a lightweight module, \eg Frozen \cite{tsimpoukelli2021multimodal}, Flamingo \cite{alayrac2022flamingo}, LLaVA \cite{liu2023visual}, Otter \cite{li2023otter}, MiniGPT-4 \cite{zhu2023minigpt}, BLIP-2 \cite{li2023blip}, AdaptorV2 \cite{gao2023llama}, etc. For example, LLaVA \cite{liu2023visual} uses a simple projection layer to align the modalities and build a visually-finetuned multimodal model; LLaVAR \cite{zhang2023llavar} extends it onto text-heavy images. 
% Considering its simple yet effective design, our model follows the architecture of LLaVA \cite{liu2023visual}. 

\noindent
\textbf{Reasoning step-by-step}
has been studied for a long time in the multimodal field. Compositional models like neural modular networks \cite{andreas2016neural, gupta2019neural, hu2018explainable} or neural symbolic reasoning \cite{yi2018neural, mao2019neuro, li2021calibrating, li2023super} decompose complex visual questions into reasoning steps, and compose modules to solve the steps. For LLMs, chain-of-thought prompting is a simple yet effective method to boost the reasoning performance \cite{wei2022chain, wang2022self, zhang2023multimodal, zhang2022automatic, yao2023tree, kojima2022large}. Recent works like VisProg \cite{gupta2023visual} and ViperGPT \cite{suris2023vipergpt} utilize LLMs as the planner to compose domain-specific models \cite{zeng2022socratic, patil2023gorilla, yang2023mm, lu2023chameleon} and external tools or APIs \cite{parisi2022talm, khot2022decomposed, schick2023toolformer, shen2023hugginggpt, qin2023toolllm} for solving complex tasks.

\begin{figure*}[t!]
\begin{center}
    \vspace{-1.0em}
    % \fbox{\rule{0pt}{2in} \rule{.9\linewidth}{0pt}}
    \includegraphics[width=0.9 \linewidth]{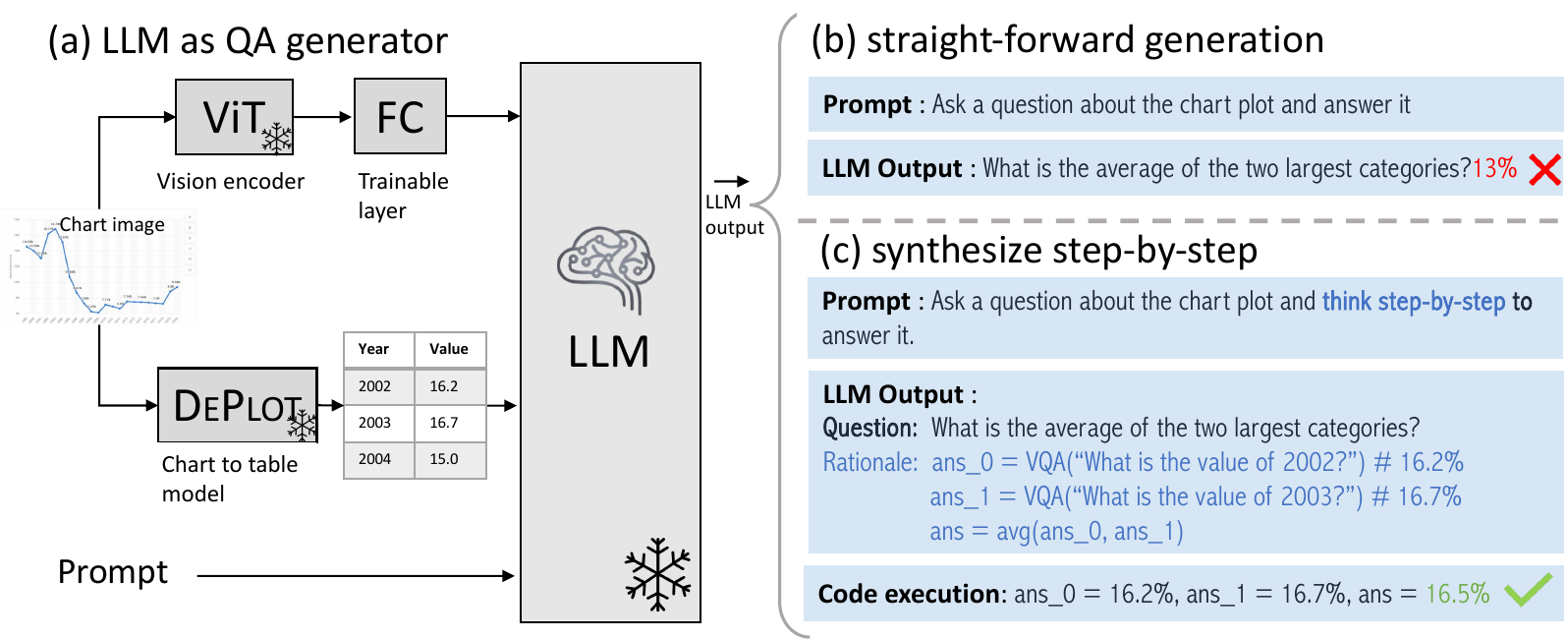}
    \vspace{-0.5em}
    \caption{Method overview. (a) The architecture of the LLM-based question-answer generator. Image features, projected by a linear layer concatenated with the predicted data table and a prompt, are fed into LLM for QA generation. (b) Straight-forward generation, where questions with answers are generated in a straightforward way. (c) Synthesizing step-by-step, which breaks the question down into rationale programs and derives the answer by executing the program.}
    \label{fig:method}
    \vspace{-1.5em}
\end{center}
\end{figure*}

\noindent
\textbf{LLMs for data generation}
has been recently studied for language tasks \cite{yu2023large, chung2023increasing, schick2021generating} or multimodal tasks \cite{liu2023visual, zhu2023minigpt}. This relates to knowledge distillation \cite{hinton2015distilling}, which teaches a smaller model by learning from a larger teacher model, benefiting semi-supervised learning \cite{chen2020big, wang2021want}. Distillation from LLMs \cite{gu2023knowledge, west2021symbolic, li2023symbolic, hsieh2023distilling, petroni2019language} can be completed by either learning from the logits or from the generated data directly. 
% Related to our work, ``distilling step-by-step'' \cite{hsieh2023distilling} suggests that smaller models learn faster provided with the rationales from LLMs; ``chain-of-thought distillation'' \cite{li2023symbolic} distills the stepwise reasoning ability from LLMs to smaller models. 
Our work is related but different from these works in that we focus on step-by-step generation, combining LLMs with templates and tools, for the chart reasoning task.

\section{Method}
% \todo{Overview of the method.}
Given a chart plot, our goal is to build a data generator that generates complex reasoning questions with answers. The generated data can be used for training the downstream chart VQA models. 

The overview of our LLM-based data generator is shown in \cref{fig:method}. In this section, we describe the data generator in a progressive manner. We first introduce an intuitive, straightforward method for data generation. Then, we introduce Synthesize Step-by-Step, which enhances the straightforward method with step-wise rationales. Finally, we introduce the settings for training the downstream chart VQA models using our generated QA data. 
% , which helps but suffer from noisy answers for complex questions

\subsection{Architecture}
The architecture of our LLM-based data generator is shown in \cref{fig:method} (a). We leverage MPT-7B \cite{team2023introducing} as the LLM for data generation. The inputs $X$ to LLM is a concatenation of three components: the projected image features $F$, the underlying data table for the chart image $T$, and a natural language prompt $P$: $$ X = [F; T; P]$$
To extract the image features, we consider the pretrained visual transformer (ViT) \cite{dosovitskiy2020image} in CLIP \cite{radford2021learning}. The ViT takes in the image patches, and extracts features for each image patch. Then we use a trainable linear layer to project the extracted image features into the input embedding space of LLM. This architecture, \ie linear layer for visual feature alignment, has been shown effective by recent multimodal large models like LLaVA \cite{liu2023visual}.

The predicted data table is critical to encode the textual information in the image. Because the pretrained ViT (in CLIP \cite{radford2021learning}) is not specifically trained for text-heavy images, the image features have limited OCR ability. Therefore, we use \deplot \cite{liu2022deplot}, which is designed for information extraction from charts, to predict the underlying data table for the given chart image. The linearized data table, which is a sequence of word tokens, is fed into the LLM.

Finally, the projected image features, the data table, and a prompt like ``Ask a question about the given chart plot and answer it'' (rephrased in several different ways), are provided as the input to LLM, for question generation.  

\begin{figure*}[t!]
\begin{center}
    \vspace{-1.0em}
    % \fbox{\rule{0pt}{2in} \rule{.9\linewidth}{0pt}}
    \includegraphics[width=0.80 \linewidth]{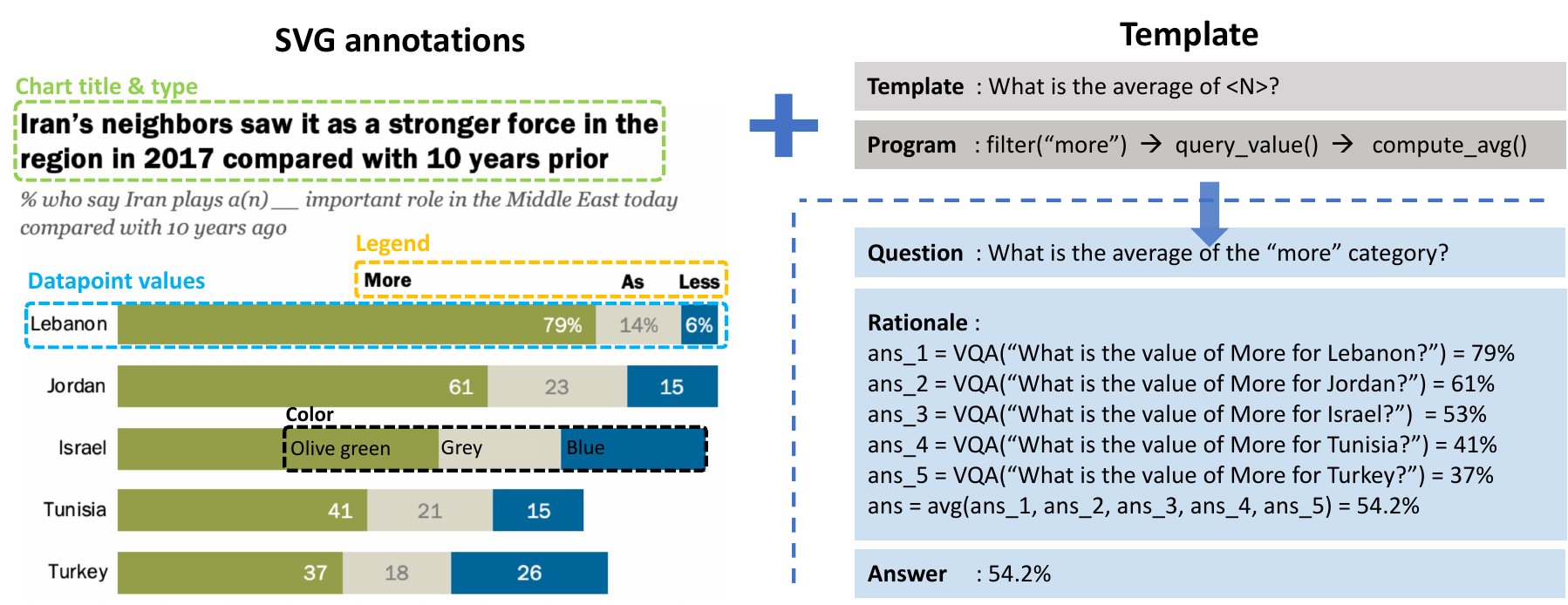}
    \vspace{-0.5em}
    \caption{Example of template-based question generation. Given a manually created \texttt{template} with defined \texttt{program}, by querying the image \texttt{SVG annotations}, templatic \texttt{questions} and \texttt{answers} can be generated, with \texttt{rationales}.}
    \label{fig:question-gen}
    \vspace{-1.5em}
\end{center}
\end{figure*}

\subsection{Straight-forward data generation}
\label{sec:straight-forward-data-generator}

Given the above inputs, a straightforward manner for data generation is to have the LLM generate question and answer directly, as shown in \cref{fig:method} (b). The output $Y$ is the generated question $Q$, followed by the corresponding answer $Q$: $$ Y = [Q; A]$$
This straightforward data generator can be trained using the existing chart VQA datasets, which contains chart images paired with question and answers. The model is trained with a next-token prediction loss. Note that the ViT and LLM are pretrained and kept frozen during training. The only trainable parameters are the linear projection layer\footnote{We also tried a two-stage training procedure, \ie finetune the whole model after training the projection layer, but did not see performance gain.}. 

As our experiments will show, this straightforward generation method yields reasonable results, generating questions and answers that are helpful for training the chart VQA models. However, the generated answers are noisy, which is improved using step-by-step generation, described next. 
% leading to an improvements ($+1\%$) in the accuracy of the downstream VQA model. 

\subsection{Synthesize step-by-step}
Here we introduce \textit{Synthesize Step-by-Step}, which shares the same model architecture as straight-forward generation, but performs data synthesizing in a step-by-step manner.

Instead of having the model generate answers directly, we train the LLM to generate them in a step-by-step manner. The LLM generates the question $Q$ with the corresponding rationales $R$. Then the answer $A$ is derived by executing the rationales using python scripts: $$Y=[Q;R], A=\mathrm{Execute}(R)$$
As the example shows in \cref{fig:method} (c), the LLM outputs the generated question and a decomposition of this complex reasoning question into a series of atom reasoning steps, \ie rationales. The rationales $R$ a step-by-step executable program, where each step in the program can be either an atom extractive VQA questions (\eg \texttt{ans\_0=VQA("What is the value of 2002?")}), or a Python-like program call for math calculation, counting, comparison, etc. (\eg \texttt{ans=avg(ans\_0, ans\_1)}). The rationale program is then parsed and executed using Python in order to derive the final answer. During execution, there are two ways to answer the atom extractive VQA questions: we can either use an off-the-shelf VQA model, or directly predict the answer using the LLM data generator itself. We experiment with both options and find the latter option to be more effective, as well as more simple, without requiring additional VQA models. 
% since empirically the LLM data generator can answer the atom questions well enough. 

\noindent
\textbf{Training with rationales.} 
To train the step-by-step generator, data \textit{with rationales} is required. This is different from the straightforward generation, which only requires question-answer pairs to train. Because existing datasets do not contain the rationale annotations for each question, we introduce a template-based QA generation pipeline, which will be described in \cref{sec:template}, to generate synthetic questions with rationales. With the template-generated data as the training corpus, we tune the generator for step-by-step generation. An additional advantage of training with template-generated data is that, in inference time, the generator can be prompted with different prompts for controllable generation. The prompts may specify the format or the type of the questions, \eg ``the question should start with how many.'', ``the question should ask about the colors in the image'', etc. A complete list of the prompts can be found in Appendix.

\noindent
\textbf{Post hoc filtering.} 
We take an additional post hoc filtering step to filter out the noisy questions. We use the the decoding score (for the whole generated sequence including question and rationales) as an indicator and filter out the questions with decoding scores lower than a threshold. Better filtering strategies, like using LLMs for data filtering, can be explored, which we will leave as future work.

\subsection{Tuning corpus: template-based QA generation}
\label{sec:template}

% \begin{figure}[b]
% \begin{center}
%     \vspace{-1.0em}
%     % \fbox{\rule{0pt}{2in} \rule{.9\linewidth}{0pt}}
%     \includegraphics[width=0.80\linewidth]{media/template_new.pdf}
%     \vspace{-0.5em}
%     \caption{Example of template-based question generation. Given a manually created \texttt{template} with defined \texttt{program}, by querying the image \texttt{SVG annotations}, templatic \texttt{questions} and \texttt{answers} can be generated, with \texttt{rationales}.}
%     \label{fig:question-gen}
%     \vspace{-1.5em}
% \end{center}
% \end{figure}

Here we describe how we use a template-based pipeline to generate questions with rationales for training the LLM-based generator. The method is motivated by the success of synthetic compositional questions in VQA datasets like CLEVR \cite{johnson2017clevr}, Super-CLEVR \cite{li2023super} and GQA \cite{hudson2019gqa}. 

\cref{fig:question-gen} illustrates the template-based question generation method. We first manually create question templates, then instantiate the templates into questions and answers by querying the rich SVG annotations of the chart image (as shown in the dotted boxes in the figure). For example, the template ``What is the average of the $\langle N \rangle$?'' can be instantiated into ``what is the average of the more category?''. The argument $\langle N \rangle$ is instantiated into ``the more category'' according to the SVG annotations, which includes the category names, values, and visual elements like colors and bounding boxes. Moreover, for each template, we define a reasoning program (in a domain-specific language) containing a series of reasoning operations (\eg \texttt{select\_color, query\_value}) that specifies the reasoning process needed to answer the question. The program is then instantiated into a step-by-step rationale.

% We created 28 templates, asking about colors, spatial relationships, counting, math calculations, etc. Each template comes with an underlying reasoning program containing a series of reasoning operations (\eg \texttt{select\_color, query\_value}). Each template is rephrased in several ways (manually) to enhance diversity. The templates and operations are listed in the Appendix.

Note that there are limitations to this template-based QA generation pipeline. The pipeline relies on the rich annotations extracted automatically from the source SVG files, and thus cannot generalize to unlabeled images. Moreover, even though some existing datasets like ChartQA provide such rich annotations crawled from websources, they are inevitably noisy, containing missing or incorrect values. In contrast, the LLM generator is more generalizable, being able to generate data for images in the wild. Therefore, in our work, these template-based questions are used as the training corpus of the LLM-based generator, which is more generalizable and scalable.

% \bhavan{We need to point out the to generate template QAs we need the details about the different elements in the plot, their color and the x and y values etc., which is available only in certain datasets when they are created or parsed. Hence this approach cannot generalize, plus such chart annotations are automatically generated during the dataset creation/parsing and hence are not accurate. Also the template approach limits the diversity of the QAs.}

\subsection{Evaluation of generated data}
To measure the usefulness of the generated QA, we use the generated data to train the chart VQA models. The accuracy improvement of chart VQA models, after training with the generated QA, is taken as the indicator of the data quality. We select \matcha \cite{liu2022matcha}, which is a transformer encoder-decoder model pretrained on large-scale text-heavy images, as our primary evaluation model, considering its modern architecture and state-of-the-art performance on standard datasets\footnote{While there are two-stage models of better performance, which extracts information from charts then prompts LLMs like Codex to answer questions, \matcha is the best single-stage model.}.
% \bhavan{Need to mention MatCha is the best performing end to end model for chart VQA datasets, hence we selected it.}

\noindent
\textbf{Summary.}
Our data generation procedure contains the following steps: first, we generate templatic QA corpus with rationales using the rich SVG annotations, which is then used to tune the LLM-based data generator; next, the LLM-based generator generates questions with step-by-step rationales; then we use Python to execute the rationales and derive the answers; finally, this LLM-augmented data, \ours, is used to train the downstream VQA model \matcha. As the experiments will show, the augmented data leads to significant performance improvements.
\section{Experiment}

\subsection{Set up}

\noindent
\textbf{Dataset.} 
Following \cite{liu2022matcha, liu2022deplot}, we run experiments on two chart VQA datasets: ChartQA and PlotQA \footnote{There are other chart VQA datasets like FigureQA, DVQA, etc., but since they are out-dated and models reach $>95\%$ accuracy, we focus on more challenging settings.}. ChartQA \cite{masry2022chartqa} contains 21k real-world chart images crawled from four web sources, with 9k human-annotated (Human split) and 23k machine-augmented (Augmented split) QA pairs. The augmented questions are mostly extractive questions, while the human-annotated questions are free-form and require complex reasoning. PlotQA \cite{methani2020plotqa} is a large synthetic dataset containing 224k chart images generated from real-world online data, with 37M synthetic questions and answers divided into two splits: V1 (8M) and V2 (29M). In addition, to show the generalizability of our LLM data generator, we generate QA for images from additional chart captioning datasets, which do not provide QAs or rich SVG annotations. We use images from three datasets: Chart-to-Text \cite{kantharaj2022chart}, VisText \cite{tang2023vistext} and ChartSumm \cite{rahman2023chartsumm}. Chart-to-Text contains 35k images from ``Statistica'' and 9k images from ``Pew''; VisText contains 9k synthetic charts of various styles; ChartSumm contains 84k images from ``Statistica'' and ``Knoema''. Note that we only use the images from the chart captioning datasets without using the captions.

% \bhavan{We should add that these datasets don't come with QA annotations and we cannot create tempalte QA from them, so we can only use our LLM based approach to generate the data, highlighting the need for LLM based approach.}

\noindent
\textbf{Generated data.}
\cref{tab: generated-stats} shows the statistics of the generated QA. To construct the template-based tuning corpus, we use 28 templates to generate 357k question-answer pairs for the ChartQA training images. The template list and the template QA can be found in the Appendix. We train our LLM data generator using the template-based corpus, and generate QAs for images in ChartQA, PlotQA, and the three chart captioning datasets. For chartQA training images, we generate 403k QAs, which reduces to 326k after step-by-step execution (dropping ones that cannot be executed) and filtering based on the decoding scores using a threshold of -10. For PlotQA training images, we generate 3M QAs, which reduces to 1.7M after execution and filtering. For the 137k chart images in the chart captioning datasets, we generate 1.6M QAs.

\begin{table}[h]
\begin{center}
\vspace{-0.5em}
\resizebox{1.0\linewidth}{!}{ 
\begin{tabular}{ll|rrr} \toprule
 & {\bf Dataset} & {\bf Images} & {\bf Generated} & {\bf Filtered} \\ \midrule
Template & ChartQA & 18,317 & 356,606 & - \\ \hline
\multirow{3}{*}{\ours} & ChartQA & 18,317 & 402,974 & 326,160 \\
 & PlotQA & 157,070 & 3,455,539 & 1,726,822 \\
 & ChartCap & 137,281 & 2,987,718 & 1,635,364 \\ \bottomrule
\end{tabular}}
\end{center}
\vspace{-1.5em}
\caption{Statistics of generated data, before and after filtering.}
\label{tab: generated-stats}
\vspace{-1.0em}
\end{table}

\begin{table*}[t]
\begin{center}
\vspace{-0.5em}
\resizebox{0.85\linewidth}{!}{ 
\begin{tabular}{ll|ccc|ccc}
\toprule
 & & \multicolumn{3}{c|}{\textbf{Accuracy}} & \multicolumn{3}{c}{\textbf{Relaxed Accuracy}} \\
 & & \textbf{avg} & \textbf{human} & \textbf{augment} & \textbf{avg} & \textbf{human} & \textbf{augment} \\ \midrule
0 & baseline (\matcha \cite{liu2022matcha}) & 47.76 & 30.21 & 65.31 & 58.54 & 39.58 & 77.50 \\ \hline
1 & syn$_{a}$ & 48.80 & 28.65 & 68.96 & 59.43 & 38.65 & 80.21 \\
2 & syn$_{at}$ w/ VQA answer & 47.76 & 28.44 & 67.08 & 57.97 & 37.81 & 78.12 \\
3 & syn$_{at}$ w/ step-VQA answer & 52.55 & 37.29 & 67.81 & 64.38 & 48.85 & 79.90 \\
4 & syn$_{at}$ w/ syn$_{at}$ answer & 53.75 & 37.92 & 69.58 & 65.10 & 48.85 & 81.35 \\
5 & syn$_{at}$ w/ step-syn$_{at}$ answer & 55.21 & 40.21 & 70.21 & 66.41 & 51.35 & 81.46 \\ \hline
% 6 & + tmpl & 56.41 & 41.15 & 71.67 & 67.66 & 52.81 & 82.50 \\
6 & syn$_{at}$ w/ step-syn$_{at}$ answer + tmpl. & 57.34 & 42.60 & 72.08 & 67.86 & 53.12 & 82.60 \\
7 & syn$_{at}$ w/ step-syn$_{at}$ answer + tmpl. + addi. & 58.65 & 45.62 & 71.67 & 69.84 & 56.56 & 83.12 \\ \bottomrule
\end{tabular}}
\end{center}
\vspace{-1.5em}
\caption{Main results on ChartQA val split. The data generator trained on \textbf{A}nnotated questions is denoted as syn$_{a}$; the data generator trained on both \textbf{A}nnotated questions and our \textbf{T}emplate-generated questions syn$_{at}$ is denoted as syn$_{at}$. Step-by-step generation is better than straight-forward generation (row 1-5). Training with a combination of step-by-step generated QA, template-generated QA (+tmpl.), and additional images from chart captioning datasets (+addi.), leads to the best model. See \cref{sec: main-results} for more details.}
\label{tab: main-results}
\vspace{-1.0em}
\end{table*}

\noindent
\textbf{Implementation details.}
Using the 357k template QA, we tune the data generator for 5 epochs with a batch size of 128. We use a cosine learning rate schedule with a max learning rate of 2e-3. The training takes about 12 hours using 8 A100 GPUs. The trainable projection layer is initialized with the first stage checkpoint of LLaVA \cite{liu2023visual}. We select the best checkpoint at 14k iterations as our best generator. For \matcha training, we take the pretrained checkpoints (base size) from HuggingFace\footnote{\url{https://huggingface.co/google/matcha-base}} and finetune the models with a combination of our generated questions and the original ChartQA or PlotQA training questions. The training setting follows \cite{lee2023pix2struct, liu2022matcha}. We use a cosine scheduler with 500 warm-up steps and a max learning rate of 0.0001. The batch size is 256. Without special description, the models are trained for 10k iterations for ChartQA (combining human and augmented splits) and 20k iterations for PlotQA because it is much larger (V1 and V2 are trained separately). We set the max number of image patches as 4096. The ChartQA training takes 3 days on 8 32GB V100 GPUs. Considering the training time, in some experiments, we reduce the image patch number to 1024 for quick experiments in 12 hours with the same setting. 

% Moreover, because the larger amount of generated training data, we train the models for longer (20k iterations for ChartQA and 30k iterations for PlotQA), which will be described in more details.

\noindent
\textbf{Evaluation metrics.} 
We suggest that both strict accuracy and relaxed accuracy should be reported, especially for the ChartQA dataset. While existing works \cite{liu2022matcha, liu2022deplot, masry2022chartqa, methani2020plotqa} evaluate answers are evaluated using relaxed accuracy, which allows 5\% error for numerical answers, we find that this relaxed accuracy metric is problematic for ChartQA. For example, many questions in ChartQA ask about years (\eg 1996, 2012), for which allowing 5\% numerical errors (\ie around 100 years) will result in incorrectly high results.

\subsection{Main results}
\label{sec: main-results}

We study our data generation method with different settings on the ChartQA dataset, with \matcha as our baseline. We first generate QA data using our method, and then use the generated data to train the \matcha model. The strict accuracy and relaxed accuracy reported on the ChartQA validation split are shown in \cref{tab: main-results}.

% We train the model with the ChartQA dataset \cite{masry2022chartqa}, which is the best chart VQA dataset contraining real-world chart plots with both human-written questions and machine-augmented questions. We train the projection layer using questions/answers from the training split, then use the trained model as a data augmentor to generate more question-answer pairs for the training chart images. 

The baseline model \matcha (row-0) achieves 47.76\% accuracy and 58.54\% relaxed accuracy. For this baseline, we finetune the pretrained \matcha checkpoint for 10k iterations with the ChartQA training split (combing human and augmented questions), using 1024 image patches. This setting follows \cite{liu2022matcha}, except that a smaller number of image patches are used for faster experiments and less GPU memory requirement. We validate the correctness of our implementation by showing that the model reaches 63.7\% relaxed accuracy using the original resolution, which matches the numbers reported in \cite{liu2022matcha} (as shown in \cref{tab: chartqa-sota}).

Row-1 shows the results for the straight-forward data generation, where we train the data generator with QAs from the ChartQA training split and use the trained generator to directly generate questions and answers. This straightforward data generation improves the accuracy by 1\% (from 47.76\% to 48.80\%), suggesting that LLMs can be helpful in data augmentation for chart reasoning.

Row-2 to row-5 shows the advantage of synthesizing step-by-step over straight-forward generation. We train the data generator with our template-generated QA, but the answers are generated with different methods. In row-2, we feed in the generated complex question into an additional VQA model (\ie baseline \matcha) to get the answers; in row-3, we generate the step-by-step rationales, and the step-wise questions are fed into the VQA model to get step-wise answers, then the final answer is derived using the step-wise answers. As can be seen, the step-wise answer generation (52.55\%) is significantly better than the direct answer generation (47.76\%), which suggests the importance of decomposing the complex questions into atom sub-questions. Row-4 and row-5 follow a similar setting, except that we use the LLM data generator to generate the answers instead of the additional VQA model. Two findings can be drawn: (a) LLM-generated answers are better than VQA-predicted answers, possibly because the LLM data generator has access to the predicted data tables as inputs; (b) step-by-step synthesizing outperforms direct generation, for both VQA-predicted and LLM-generated answers.

\begin{table*}[t]
\begin{center}
\vspace{-1.0em}
\resizebox{.83\linewidth}{!}{ 
\begin{tabular}{lr|ccc|ccc}
\toprule
 & & \multicolumn{3}{c|}{\textbf{Accuracy}} & \multicolumn{3}{c}{\textbf{Relaxed Accuracy}} \\
 & \textbf{params} & \textbf{avg} & \textbf{human} & \textbf{augment} & \textbf{avg} & \textbf{human} & \textbf{augment} \\ \midrule
VL-T5-OCR \cite{masry2022chartqa} & - & - & - & - & 41.56 & - & - \\
VisionTapas-OCR \cite{masry2022chartqa} & - & - & - & - & 45.52 & - & - \\
% PaLI-17B (res. 588) & - & - & - & 47.6 & 30.4 & 64.9 \\
Pix2Struct$_{4096}$ \cite{lee2023pix2struct} & 282M & - & - & - & 56.0 & 30.5 & 81.6 \\
\textsc{MatCha}$_{4096}$ \cite{liu2022matcha} & 282M & - & - & - & 64.2 & 38.2 & 90.2 \\
\textsc{DePlot}+FlanPaLM+Codex \cite{liu2022deplot} & 540B & - & - & - & 79.3 & 67.6 & 91.0 \\ 
PaLI-X \cite{chen2023pali} & 55B & - & - & - & 72.3 & - & - \\ \hline
Pix2Struct$_{4096}$ (reimpl.) & 282M & 50.44 & 24.64 & {\bf 76.24} & 58.24 & 32.00 & {\bf 84.48} \\
Pix2Struct$_{1024}$ + \ours & 282M & 51.77 & 34.69 & 68.85 & 63.02 & 46.25 & 79.79 \\
Pix2Struct$_{4096}$ + \ours & 282M & {\bf 55.83} & {\bf 40.10} & 71.56 & {\bf 66.82} & {\bf 51.67} & 81.98  \\ \hline
\textsc{MatCha}$_{4096}$ (reimpl.) & 282M  & 55.68 & 31.04 & 80.32 & 63.72 & 37.76 & 89.68 \\
\textsc{MatCha}$_{1024}$ + \ours & 282M & 60.28 & 41.92 & 78.64 & 69.88 & 51.76 & 88.00 \\
{\bf \textsc{MatCha}$_{4096}$ + \ours} & 282M & {\bf 63.36} & {\bf 44.88} & {\bf 81.84} & {\bf 72.64} & {\bf 53.92} & {\bf 91.36} \\ \bottomrule
\end{tabular}}
\end{center}
\vspace{-1.5em}
\caption{Comparison with SoTAs on ChartQA test split. Subscripts ($_{1024}$, $_{4096}$) means image token numbers (resolution). With our generated data, both Pix2Struct and \matcha achieve significantly better performance, even comparable to much larger models.} 
\label{tab: chartqa-sota}
\vspace{-1.0em}
\end{table*}

% \bhavan{We should also add a column to show model sizes to highlight it's not fair to compare with very large models like GPT-4 kind}

In row-6, we combine the step-by-step generated data with the templated-generated QAs and show that training with both leads to further better performance (57.34\%). In row-7, we generate QAs for the chart images from additional chart captioning datasets. The final model, trained with a combination of all the synthetic data, gets 58.65\% accuracy. Compared with the baseline, training with the augmented data leads to a 10.89\% accuracy improvement. Moreover, the improvement is more significant on the human-written questions ($+15.41\%$), which requires complex reasoning, than the augmented extractive questions ($+6.36\%$). The results suggest that our synthetic data improves both the extractive ability and the reasoning ability, with the latter being more significant. We show examples in the Appendix.

\subsection{Comparison with SoTA}

To compare our method with the best performing models, we apply our generated data onto two models, Pix2Struct \cite{lee2023pix2struct} and \matcha \cite{liu2022matcha}, and report the results on the ChartQA test split in \cref{tab: chartqa-sota}. For faster training, we first train the model (for both Pix2struct and \matcha) with a max of 1024 image patches using our best setting (\ie row-7 in \cref{tab: main-results}), then finetune with 4096 image patches to get the final result. We report both the 1024-resolution results and the 4096-resolution results. 

As can be seen, our generated data improves the overall relaxed accuracy of \pixtostruct by 8.58\% and \matcha by 8.92\%. Moreover, the improvement is much more significant on human-written questions, \eg the relaxed accuracy of \matcha on human-written questions improves from 38\% to 54\%. Interestingly, with our augmented data, the low-resolution (1024) models can already outperform their high-resolution (4096) counterparts without data augmentation. The significant performance gain indicates that learning with synthetic data is extremely helpful for better reasoning ability with human-written queries.

Our best performing \matcha achieves 72.6\% relaxed accuracy, which beats all existing end-to-end chart reasoning models. Note that \textsc{DePlot}+FlanPaLM+Codex is a two-stage model that first extracts information from the charts, then prompts external LLMs like FlanPalm (540B) Codex (12B) for question answering. The method relies on much larger models of billions of parameters and thus is not comparable with our method, which is much smaller and not rely on external LLMs for inference. 

% Our best \matcha model, without relying on external LLMs during inference, achieve the best performance. \bhavan{Need to specify these LLM is billions of parameters and trained on insane amount of data, compared to the LLM we use, and also it's not an end to end model. Otherwise reviewers would point it out.}

% \cref{tab: chartqa-sota} compares our method on the ChartQA test split with the best performing models, including Pix2Struct \cite{lee2023pix2struct}, \matcha \cite{liu2022matcha} and \deplot with Codex \cite{liu2022deplot}. 

\subsection{Generalization}

To show the generalization ability of the proposed data generation method, we run experiments on the PlotQA dataset. We directly use our data generator (trained on template questions on ChartQA) to generate QA for the PlotQA training images. The results are shown in \cref{tab: plotqa-sota}. As can be seen, our method, when applied to \matcha, achieves state-of-the-art performance (92.89\% relaxed accuracy) on PlotQA. Despite the synthetic nature of this dataset and the saturated model performance, our augmented data can still help on both low-resolution ($+3.46\%$) and high-resolution ($+0.92\%$) settings.

\begin{table}[b]
\begin{center}
\vspace{-1.5em}
\resizebox{.95\linewidth}{!}{  
\begin{tabular}{l|ccc}
\toprule
 & \multicolumn{3}{c}{\textbf{Relaxed Accuracy}} \\
 & \textbf{avg} & \textbf{V1} & \textbf{V2} \\ \midrule
VisionTapas-OCR \cite{masry2022chartqa} & 53.90 & 65.30 & 42.50 \\
% PaLI-17B (res. 588) & - & - & - & - & 39.8 & 64.5 & 15.2 \\
VL-T5-OCR \cite{masry2022chartqa} & 65.96 & 75.90 & 56.02 \\
\textsc{DePlot}+FlanPaLM+Codex \cite{liu2022deplot} & 66.6	& 62.2 & 71.0 \\
Pix2Struct \cite{lee2023pix2struct} & 72.5 & 73.2 & 71.9 \\
\textsc{MatCha}$_{4096}$ \cite{liu2022matcha}& {\bf 91.5} & {\bf 92.3} & {\bf 90.7} \\ \hline
\textsc{MatCha}$_{1024}$ (reimpl.) & 74.95 & 73.88 & 76.02 \\
\textsc{MatCha}$_{1024}$ + \ours & 78.41 & 78.44 & 78.38 \\ \hline
\textsc{MatCha}$_{4096}$ (reimpl.) & 91.97 & 92.64 & 91.30 \\
% ours(20k) & 4096 & 50.31 & 74.96 & 25.66 & 91.23 & 91.74 & 90.72 \\
\textsc{MatCha}$_{4096}$ + \ours & {\bf 92.89} & {\bf 93.94} & {\bf 91.84} \\ \bottomrule
\end{tabular}}
\end{center}
\vspace{-1.5em}
\caption{Comparison with SoTAs on PlotQA test split. With our generated data, \matcha achieves the SoTA performance.}
\label{tab: plotqa-sota}
\vspace{-1.5em}
\end{table}

\subsection{Ablations and analysis}

\begin{figure*}[t]
\begin{center}
    \vspace{-1.0em}
    % \fbox{\rule{0pt}{2in} \rule{.9\linewidth}{0pt}}
    \includegraphics[width=0.80\linewidth]{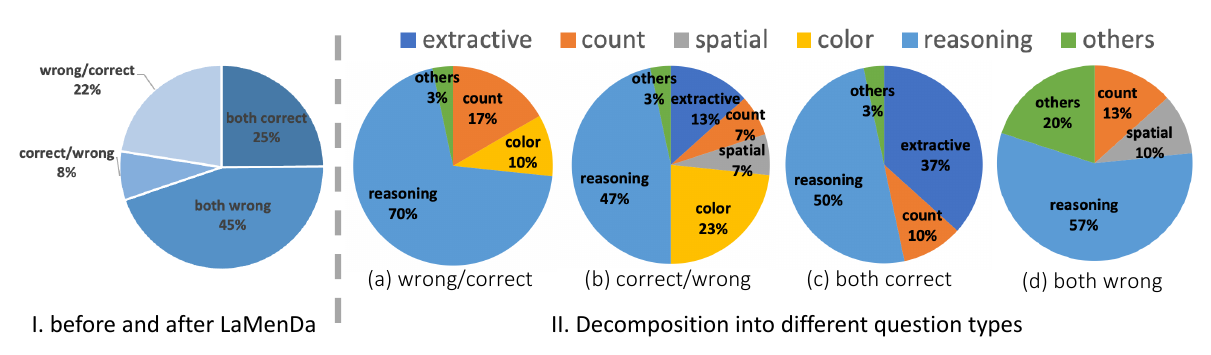}
    \vspace{-0.5em}
    \caption{Analysis of model predictions \textbf{before} and \textbf{after} training with \ours. In I (left), ChartQA val questions are divided into four categories according to the model prediction correctness (before/after). II (right) further decomposes each category into different question types. \ours improves the model prediction, and the improvement is most significant on \textit{reasoning} questions.}
    \label{fig:failure_stats}
    \vspace{-2.0em}
\end{center}
\end{figure*}

\begin{figure}[b]
\begin{center}
    \vspace{-1.0em}
    % \fbox{\rule{0pt}{2in} \rule{.9\linewidth}{0pt}}
    \includegraphics[width=0.70\linewidth]{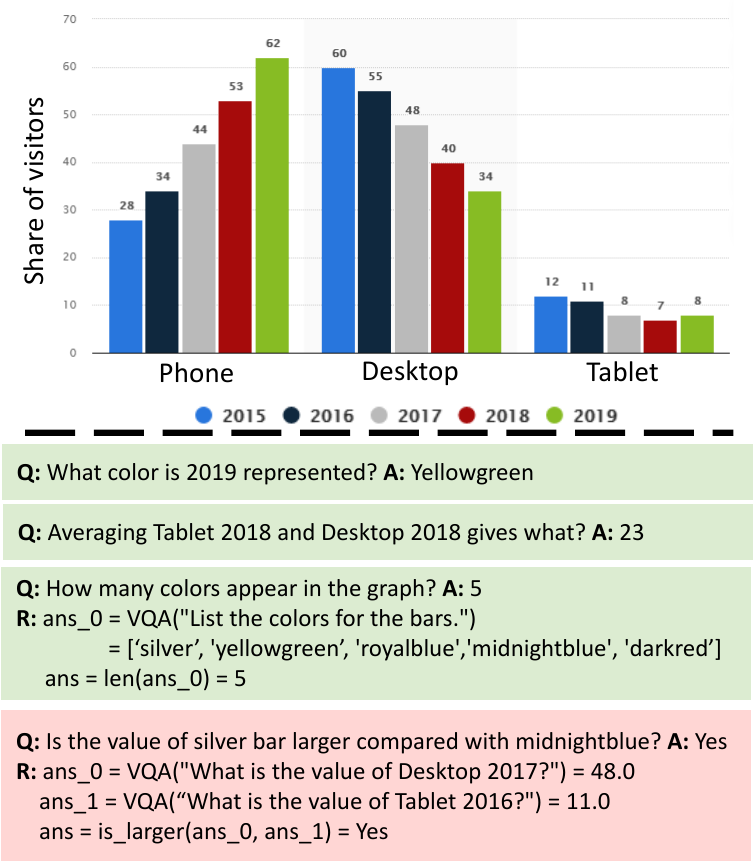}
    \vspace{-1.0em}
    \caption{Examples of generated QA, 3 good examples and 1 failure case. Rationales are only shown for 2 examples for brevity.}
    \vspace{-1.5em}
    \label{fig:example}
\end{center}
\end{figure}

\noindent
\textbf{Qualitative analysis and failure cases.}
\cref{fig:example} shows examples of our LLM-generated QA, including good and bad examples. Most generated questions are specific to the given image, asking about the categories plotted, such as ``2019'' or ``Tablet 2018'', indicating that the model is able to utilize the information of the input. Moreover, various questions are generated, including counting, colors, average, etc. In the failure example, the question requires comparing the ``silver bar'' with the ``midnightblue'' bar, with reasonable rationales. However, there are multiple bars in the specified colors in the image, resulting in ambiguous/invalid questions. When looking at more failure cases, we find that there are several typical failure modes. For example, questions are sometimes invalid/ambiguous (\eg asking about the top/bottom bar for a vertical bar plot); questions can be hallucinated or irrelevant (\eg asking about a category that is not shown in the figure); rationales and answers are not correct, etc., which opens up opportunities for future works.

\noindent
\textbf{Before and after \ours.}
To gain insights on how \ours helps model training, we conduct a comparison analysis of \matcha \textbf{``before'' and ``after''} training with our LLM-generated data, \ie the baseline model versus our best model. According to the predictions of these two models, we divide the 960 human-written questions from ChartQA val split into 4 groups: both model answers correctly, both models answers wrongly, the before model answers correctly, but the after model answers wrongly, and vice versa. The statistics of these four groups are shown in \cref{fig:failure_stats}. From \cref{fig:failure_stats} I. (left side), we see that \ours is helpful for 22\% of the questions, while there are still 45\% of the questions where both models cannot answer correctly. In \cref{fig:failure_stats} II. (right side), we randomly sample 30 questions from each of the four groups and manually classify them into six question types: extractive, count, spatial, color, reasoning, and others. As shown in (a), \ie the questions where the ``before'' model answers wrongly and our model answers correctly, most (70\%) of the questions belong to the reasoning category, indicating that our data helps most on the reasoning questions. Our data also helps with color questions (10\%) and count questions (17\%). However, our data does not help with extractive questions (0\%). In addition, comparing the distribution of (c) both correct with (d) both wrong, we can see that extractive questions are easy for the models, while reasoning questions are very challenging.

In summary, \cref{fig:failure_stats} shows the effectiveness of our data for the \textbf{reasoning questions} over other question types. However, reasoning questions are still a big challenge for current models, which requires further exploration.

\noindent
\textbf{Different question types.}
In \cref{tab: llm-pertype}, we study the improvements brought by each type of the generated questions. The total 326k generated QAs are grouped into 7 different groups, including questions about color understanding, spatial reasoning, counting, minimum/maximum reasoning, average, comparison (\eg larger than, smaller than), and other math calculations. This grouping is done by different prompting of the data generation LLM, as specified in the Appendix. The improvements from different question types are distinct and can be accumulated. As we see, each of the seven types improves the model performance by less than $3\%$, while combining all of them leads to $7.03\%$ improvement. Moreover, among the seven types, the ``math calculation'' questions brings the most improvements ($2.76\%$).

\begin{table}[h]
\begin{center}
\vspace{-0.5em}
\resizebox{.70\linewidth}{!}{ 
\begin{tabular}{l|r|ccc}
\toprule
 & \multicolumn{1}{|c|}{\bf \#QA} & \textbf{avg} & \textbf{human} & \textbf{aug} \\ \midrule
baseline & 28,299 & 47.76 & 30.21 & 65.31 \\ \hline
+colors & +46,512 & 47.81 & 29.90 & 65.73 \\
+spatial & +49,387 & 49.27 & 30.63 & 67.92 \\
+count & +36,705 & 47.40 & 29.17 & 65.63 \\
+minmax & +71,788 & 49.58 & 31.87 & 67.29 \\
+average & +50,717 & 48.96 & 31.35 & 66.56 \\
+compare & +14,690 & 47.66 & 28.85 & 66.46 \\
+calculation & +56,361 & 50.52 & 33.85 & 67.19 \\ \hline
+all & 326,160 & 54.79 & 39.27 & 70.31 \\ \bottomrule
\end{tabular}}
\end{center}
\vspace{-1.5em}
\caption{A detailed look at the effect for different question types. Strict accuracy on ChartQA val split is shown. Each type helps and combining all of them leads to a further performance gain.}
\label{tab: llm-pertype}
\vspace{-1.5em}
\end{table}

\section{Conclusion}

We propose Synthesize Step-by-Step, which utilizes LLMs to augment the training data for the chart reasoning task. We show that the step-by-step generation, which decomposes the complex reasoning questions into easier sub-questions, is critical for synthesizing good-quality data. We run experiments on two standard benchmarks and show significant improvements by training with the augmented data. Specifically, the relaxed accuracy on the human-written questions is improved from 37.8\% to 53.9\% on ChartQA. The large improvement indicates the promise of boosting the model's reasoning ability by data generation using LLMs, from which the strong step-by-step reasoning ability can be distilled into downstream models.

% \textbf{Limitations.}

%%%%%%%%% REFERENCES
% \newpage
{
    \small
    \bibliographystyle{ieee_fullname}
    \bibliography{references}
}

\newpage
\appendix
\onecolumn

\section{Data scaling analysis}

To analyze the influence of data amount, we train \matcha using different amount (25\%, 50\%, 75\%, 100\%) of our \ours data, and report results in the \cref{tab: data-scaling}. We use the same setting and data as row-5 in Tab-2, except that the training schedule is shorter (5k iterations) due to time limitations. As shown, improvements of our generated data follow a typical data scaling law: improvements are large initially (at 25\%) then continues with smaller margins. 

\begin{table}[h!]
\begin{center}
% \resizebox{0.82\linewidth}{!}{ 
\begin{tabular}{lccccc} \toprule
\textbf{data amount}      & \textbf{0\%} & \textbf{25\%} & \textbf{50\%} & \textbf{75\%} & \textbf{100\%} \\ \midrule
\textbf{human}     & 30.21             &   34.38            & 35.73              & 36.25              & 37.19                \\
\textbf{augment}   & 65.31             &   66.87            & 68.85              & 69.69              & 68.96               \\ \midrule
\textbf{avg.} & 47.76             & 50.63         & 52.29         & 52.97         & 53.07          \\ \bottomrule
\end{tabular} 
% }
\end{center}
\vspace{-1.5em}
\caption{Results with different amount of generated data.}
\label{tab: data-scaling}
\end{table}

\section{Template-based QA generation}

\textbf{Results using template-generated data only.}
We report the result using template-generated QA data only, under the same setting as Tab-2 (1024 patches, on ChartQA): training with only template QAs leads to 54.32\% strict accuracy (38.23\% on human, 70.73\% on augmented). This result is better than baseline (47.76\%), but lower than LLM generator which gets 58.65\%. Additionally, although template QAs are very clean, they (a) are not free-form and (b) rely on groundtruth SVG metadata, thus cannot generalize to images without SVG, which highlights the advantages of LLM generator.

\noindent
\textbf{Template list.}
Empirically, we find that template diversity is crucial for the model performance, otherwise the LLM generator may overfit to rigid templates. For example, in the early stage of the project, our LLM generator gets only 49.8\% accuracy trained with 16 templates without rephrasing, which is much lower than the current 55.2\% (row 5 Tab-2). \cref{tab: templates} shows the final list of templates that are used for the templated-based question-answer generation pipeline. For each template, we manually define a functional program as inspired by the CLEVR dataset \cite{johnson2017clevr}. The functions include 7 basic operations like \texttt{SUM}, \texttt{COUNT}, \texttt{COMPARE}, and a \texttt{VQA} operation. The execution is motivated by VisProg \cite{gupta2023visual}, where each operation is executed by a predefined Python function.

\begin{table}[h]
\begin{center}
\begin{tabular}{l} \toprule
\multicolumn{1}{c}{\textbf{Color}} \\ \hline
1. What color is $\langle N \rangle$ represented? \\
2. What is the value of the $\langle C \rangle \langle F \rangle$? \\ 
3. Which category is represented by $\langle C \rangle$? \\ \toprule

\multicolumn{1}{c}{\textbf{Spatial}} \\ \hline
4. What is the value of the $\langle S \rangle$ bar? \\
5. What does the $\langle S \rangle$ bar represent? \\
6. What is the value of the second bar from the $\langle S \rangle$? \\
7. What is [represented by] the second bar from the $\langle S \rangle$? \\
8. What is the value of the third bar from the $\langle S \rangle$? \\
9. What is represented by the third bar from the $\langle S \rangle$? \\ \toprule

\multicolumn{1}{c}{\textbf{Count}} \\ \hline
10. How many $\langle F \rangle$s are shown in the plot? \\
11. How many $\langle C \rangle$ bars are shown in the plot? \\
12. How many colors are used to represent the $\langle F \rangle$s in the plot? \\ \toprule

\multicolumn{1}{c}{\textbf{Math}} \\ \hline
13. What is the average of $\langle N \rangle$? \\
14. What is the max [value] of $\langle N \rangle$? \\
15. What is the min value of the $\langle N \rangle$? \\
16. What is the [total] sum [value] of $\langle L \rangle$ and $\langle L2 \rangle$? \\
17. What is the difference between [values of] $\langle L \rangle$ and $\langle L2 \rangle$? \\
18. What is the value of the smallest category[in the chart]? \\
19. What is the value of the largest category[ in the chart]? \\
20. What is the smallest category? \\
21. What is the largest category? \\
22. What is the average [value] of the two smallest categories[ in the chart]? \\
23. What is the average [value] of the two largest categories[ in the chart]? \\
24. What is the difference between the largest [category] and the smallest category? \\
25. What is the ratio [value] of $\langle L \rangle$ and $\langle L2 \rangle$? \\
26. How many times [is] $\langle L \rangle$ bigger than $\langle L2 \rangle$? \\
27. What is the average of $\langle L \rangle$ and $\langle L2 \rangle$? \\
28. Is [the value of] $\langle L \rangle$ more than $\langle L2 \rangle$? \\ \bottomrule
\end{tabular}
\vspace{-1.5em}
\end{center}
\caption{List of templates.}
\label{tab: templates}
\end{table}

\section{Prompts}
\cref{tab: prompts} shows the prompts to prompt the LLM-based data generator, for controllably generating questions and answers.

\begin{table}[h]
\begin{center}
\begin{tabular}{l} \toprule
1. The question should be similar to this: ... \\
2. The question should be free form. \\
3. The question should require color understanding of the image. \\
4. The question should require counting. \\
5. The question should require counting of colors. \\
6. The question should require counting and color understanding. \\
7. The question should require spatial understanding of the image. \\
8. The question should require math reasoning about min. \\
9. The question should require math reasoning to compute min. \\
10. The question should require math reasoning to compute average of two categories. \\
11. The question should require math reasoning to compute average. \\
12. The question should require math reasoning to compute max. \\
13. The question should require math reasoning about the difference between max and min. \\
14. The question should require math reasoning to compute difference. \\
15. The question should require math reasoning about comparison. \\
16. The question should require math reasoning about average and max. \\
17. The question should require math reasoning to compute sum. \\
18. The question should require math reasoning about max. \\
19. The question should require math reasoning about average and min. \\
20. The question should require math reasoning to compute ratio. \\
21. The question should require color understanding and math reasoning to compute difference. \\
22. The question should require color understanding and math reasoning about comparison. \\
23. The question should require spatial understanding and math reasoning to compute difference. \\
24. The question should require spatial understanding and math reasoning about average. \\ \bottomrule
\end{tabular}
\vspace{-1.5em}
\end{center}
\caption{List of prompts for prompting the LLM-based data generator.}
\label{tab: prompts}
\end{table}

\section{Dataset statistics}

\cref{tab: data-statistics} shows detailed statistics for the ChartQA and the PlotQA datasets.
\cref{tab: cap122k-stats} shows the detailed statistics of the chart captioning datasets. We generated questions and answers using our LLM-based data generator for the chart images in these chart captioning datasets. 

% \begin{table}[h]
% \begin{center}
% \begin{tabular}{l|rr}\toprule
%  & \multicolumn{1}{l}{train} & \multicolumn{1}{l}{val} \\ \midrule
% color & 24173 & 2705 \\
% spatial & 140852 & 6736 \\
% count & 22001 & 1598 \\
% math & 169580 & 9894 \\ \hline
% all & 356606 & 20933 \\ \bottomrule
% \end{tabular}
% \vspace{-1.5em}
% \end{center}
% \caption{Statistics for the template-generated QA on ChartQA.}
% \label{tab: templateQA-stats}
% \end{table}

\begin{table*}[h]
\begin{center}
\resizebox{.75\linewidth}{!}{ 
\begin{tabular}{l|ccc|ccc} \toprule
 & \multicolumn{3}{c|}{\textbf{ChartQA}} & \multicolumn{3}{c}{\textbf{PlotQA}} \\ 
 & \textbf{Images} & \textbf{HumanQA} & \textbf{AugmentedQA} & \textbf{Images} & \textbf{V1 QA} & \textbf{V2 QA} \\ \midrule
train & 18,317 & 7,398 & 20,901& 157,070 & 5,733,893 & 20,249,479 \\
val & 1,056 & 960 & 960 & 33,653 & 1,228,468 & 4,360,648 \\
test & 1,509 & 1,250 & 1,250 & 33,660 & 1,228,313 & 4,342,514 \\ \bottomrule
\end{tabular}}
\end{center}
\vspace{-1.5em}
\caption{Statistics for ChartQA and PlotQA.}
\label{tab: data-statistics}
% \vspace{-1.0em}
\end{table*}

\begin{table}[h]
\begin{center}
\begin{tabular}{cl|rrrr} \toprule
 &  & \#images & \#captions & \#llava\_pred & \#filter-10 \\ \midrule
\multirow{4}{*}{Chart-to-text \cite{kantharaj2022chart}} & statista\_two\_col & 27868 & 27868 & 603283  & 613096 \\
 & statista\_multi\_col & 6943 & 6943 & 152746  & 107752 \\
 & pew\_two\_col & 1486 & 1486 & 31806 & 23521 \\
 & pew\_multi\_col & 7799 & 7799 & 171578 & 113967 \\ \hline
VisText \cite{tang2023vistext} & vistext & 8822 & 9969 & 172319 & 76788 \\ \hline
\multirow{2}{*}{ChartSumm \cite{rahman2023chartsumm}} & img\_list\_s & 40985 & 32786 & 901670 & 364218 \\
 & img\_list\_k & 43378 & 34702 & 954316 & 336022 \\ \hline
All & - & 137281 & 121553 & 2987718 & 1635364 \\ \bottomrule
\end{tabular}
\vspace{-1.5em}
\end{center}
\caption{Statistics for the chart captioning datasets.}
\label{tab: cap122k-stats}
\end{table}

% \begin{table}[h]
% \begin{center}
% \begin{tabular}{l|ccc} \toprule
%  & \textbf{avg} & \textbf{human} & \textbf{augmented} \\ \midrule
% baseline & 47.92 & 30.31 & 65.63 \\ \hline
% +colors & 49.43 & 32.40 & 66.98 \\
% +spatial & 48.85 & 30.52 & 67.71 \\
% +counting & 48.39 & 30.21 & 66.67 \\
% +math & 55.05 & 40.83 & 70.10 \\ \hline
% +all & 56.25 & 41.04 & 71.77 \\ \bottomrule
% \end{tabular}
% \vspace{-1.5em}
% \end{center}
% \caption{Accuracy (strict) for each template type. \zhuowan{I will do a similar table for our method, this one may be removed or put into appendix}}
% \label{tab: template-per-type}
% \end{table}

\section{Additional results}
\cref{tab: plotqa-sota-full} shows the full results including relaxed accuracy and strict accuracy for \cref{tab: plotqa-sota} in the main paper. \cref{tab: llm-pertype-full} shows the full results including relaxed accuracy and strict accuracy for \cref{tab: llm-pertype} in the main paper.

\begin{table*}[h]
\begin{center}
\begin{tabular}{l|ccc|ccc}
\toprule
 & \multicolumn{3}{c|}{\textbf{Accuracy}} & \multicolumn{3}{c}{\textbf{Relaxed Accuracy}} \\
 & \textbf{avg} & \textbf{V1} & \textbf{V2} & \textbf{avg} & \textbf{V1} & \textbf{V2} \\ \midrule
VisionTapas-OCR \cite{masry2022chartqa} & - & - & - & 53.90 & 65.30 & 42.50 \\
% PaLI-17B (res. 588) & - & - & - & - & 39.8 & 64.5 & 15.2 \\
VL-T5-OCR \cite{masry2022chartqa} & - & - & - & 65.96 & 75.90 & 56.02 \\
\textsc{DePlot}+FlanPaLM(540B)+Codex \cite{liu2022deplot} & - & - & - & 66.6	& 62.2 & 71.0 \\
Pix2Struct \cite{lee2023pix2struct} & - & - & - &  72.5 & 73.2 & 71.9 \\
\matcha \cite{liu2022matcha} & - & - & - & {\bf 91.5} & {\bf 92.3} & {\bf 90.7} \\ \hline
\textsc{MatCha}$_{1024}$ (reimpl.) & 41.58 & 66.66 & 16.50 & 74.95 & 73.88 & 76.02 \\
\textsc{MatCha}$_{1024}$ + \ours & 43.04 & 68.48 & 17.60 & 78.41 & 78.44 & 78.38 \\ \hline
\textsc{MatCha}$_{4096}$ (reimpl.) & 50.89 & 76.14 & 25.64 & 91.97 & 92.64 & 91.30 \\
% ours(20k) & 4096 & 50.31 & 74.96 & 25.66 & 91.23 & 91.74 & 90.72 \\
\textsc{MatCha}$_{4096}$ + \ours & {\bf 51.40} & {\bf 76.42} & {\bf 26.38} & {\bf 92.89} & {\bf 93.94} & {\bf 91.84} \\ \bottomrule
\end{tabular}
\end{center}
\vspace{-1.5em}
\caption{Full results for \cref{tab: plotqa-sota}: comparison with SoTAs on PlotQA test split. With our generated data, \matcha achieves the SoTA performance.}
\label{tab: plotqa-sota-full}
\vspace{-1.0em}
\end{table*}

\begin{table*}[h]
\begin{center}
\begin{tabular}{l|c|ccc|ccc}
\toprule
 & \multicolumn{1}{|c|}{\multirow{2}{*}{\textbf{\# Questions}}} & \multicolumn{3}{c|}{\textbf{Accuracy}} & \multicolumn{3}{c}{\textbf{Relaxed Accuracy}} \\
 & \multicolumn{1}{|c|}{} & \textbf{avg} & \textbf{human} & \textbf{augment} & \textbf{avg} & \textbf{human} & \textbf{augment} \\ \midrule
baseline & 28,299 & 47.76 & 30.21 & 65.31 & 58.54 & 39.58 & 77.50 \\
+colors & +46,512 & 47.81 & 29.90 & 65.73 & 59.32 & 40.21 & 78.44 \\
+spatial & +49,387 & 49.27 & 30.63 & 67.92 & 61.09 & 43.33 & 78.85 \\
+count & +36,705 & 47.40 & 29.17 & 65.63 & 58.28 & 38.54 & 78.02 \\
+minmax & +71,788 & 49.58 & 31.87 & 67.29 & 59.38 & 40.63 & 78.13 \\
+average & +50,717 & 48.96 & 31.35 & 66.56 & 60.00 & 40.52 & 79.48 \\
+compare & +14,690 & 47.66 & 28.85 & 66.46 & 58.59 & 38.85 & 78.33 \\
+calculation & +56,361 & 50.52 & 33.85 & 67.19 & 62.40 & 45.21 & 79.58 \\ \hline
+all & +326,160 & 54.79 & 39.27 & 70.31 & 66.15 & 50.42 & 81.88 \\ \bottomrule
\end{tabular}
\end{center}
\vspace{-1.5em}
\caption{Full results for \cref{tab: llm-pertype}: a detailed look at the effect for different question types. Strict accuracy on ChartQA val split is shown. Each type helps and combining all of them leads to a further performance gain.}
\label{tab: llm-pertype-full}
\vspace{-1.0em}
\end{table*}

\section{Examples comparing \textsc{MatCha} with and without \ours}

\begin{figure}
\begin{subfigure}{.5\textwidth}
  \centering
  \includegraphics[width=.8\linewidth]{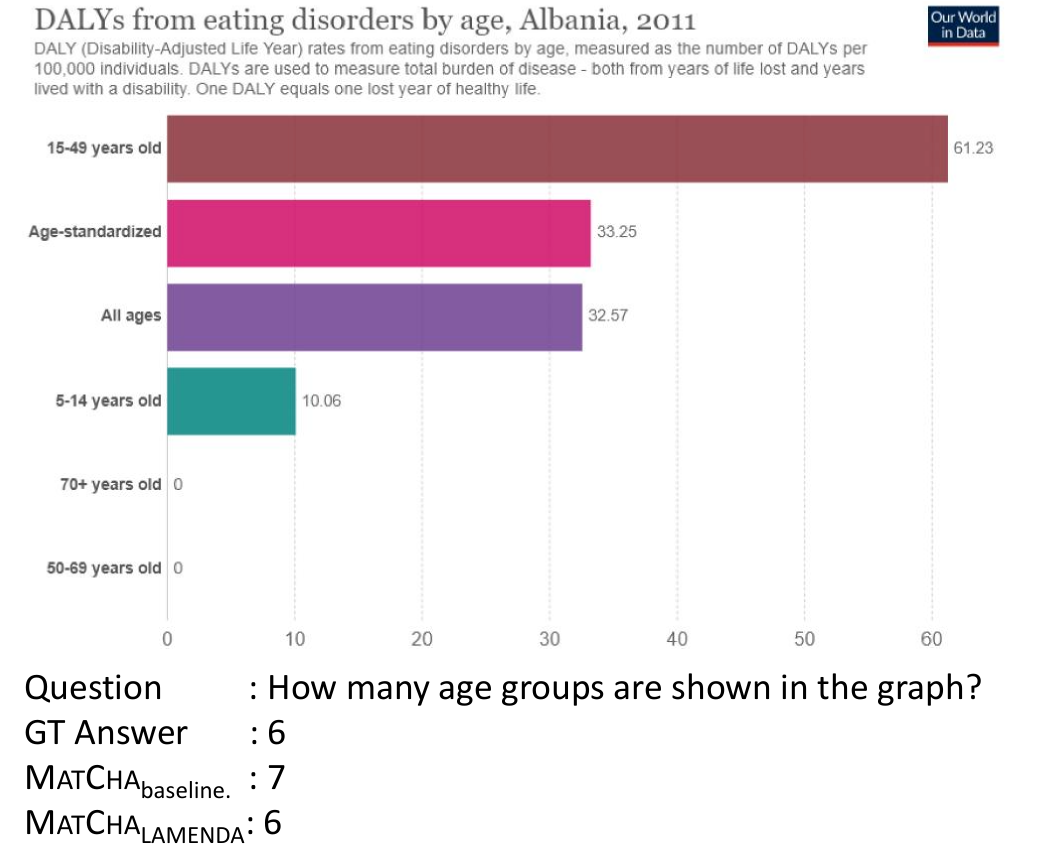}
  \caption{}
  \label{fig:sfig1}
\end{subfigure}%
\begin{subfigure}{.5\textwidth}
  \centering
  \includegraphics[width=.8\linewidth]{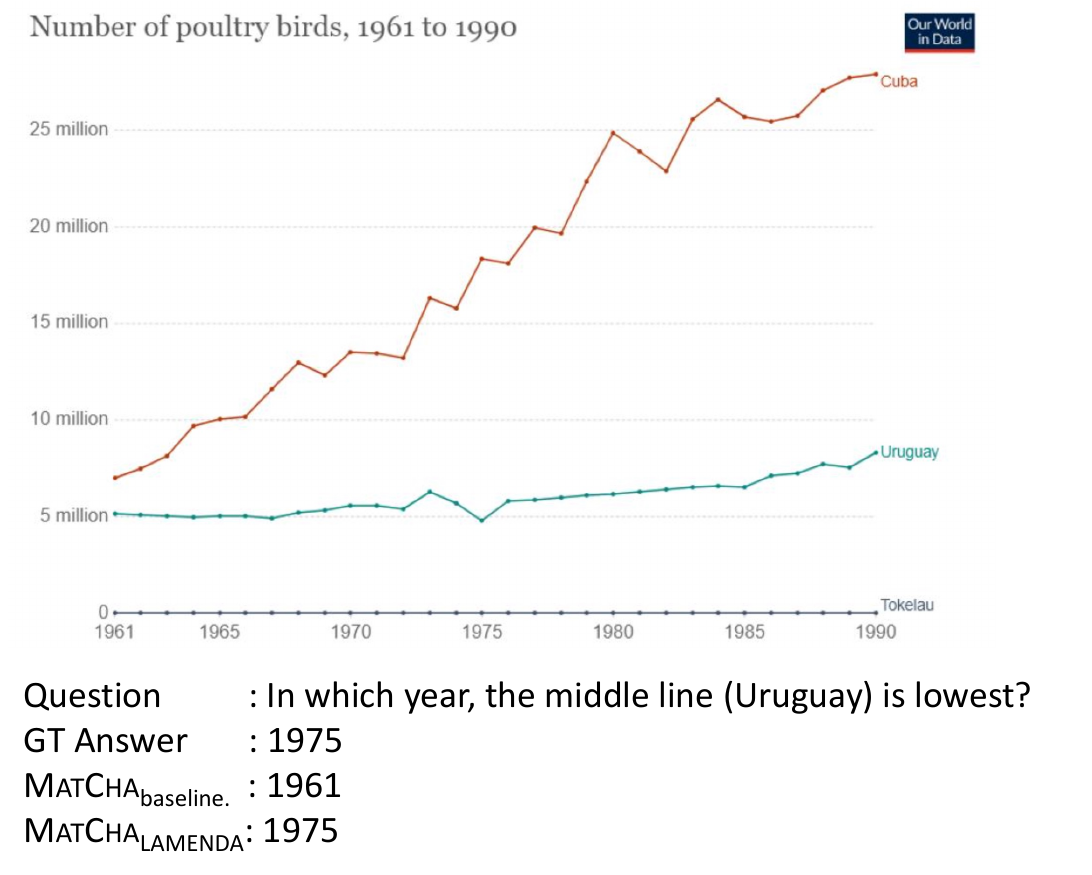}
  \caption{}
  \label{fig:sfig2}
\end{subfigure}
\begin{subfigure}{.5\textwidth}
  \centering
  \includegraphics[width=.8\linewidth]{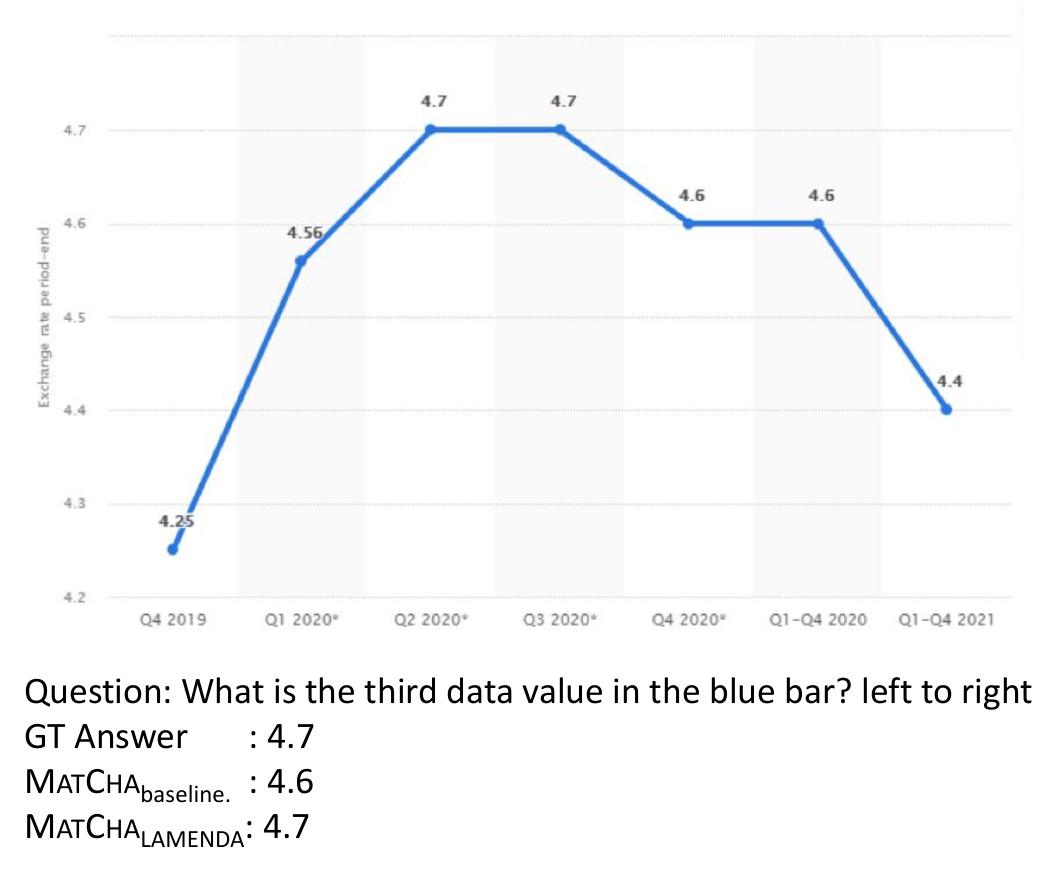}
  \caption{}
  \label{fig:sfig2}
\end{subfigure}
\begin{subfigure}{.5\textwidth}
  \centering
  \includegraphics[width=.8\linewidth]{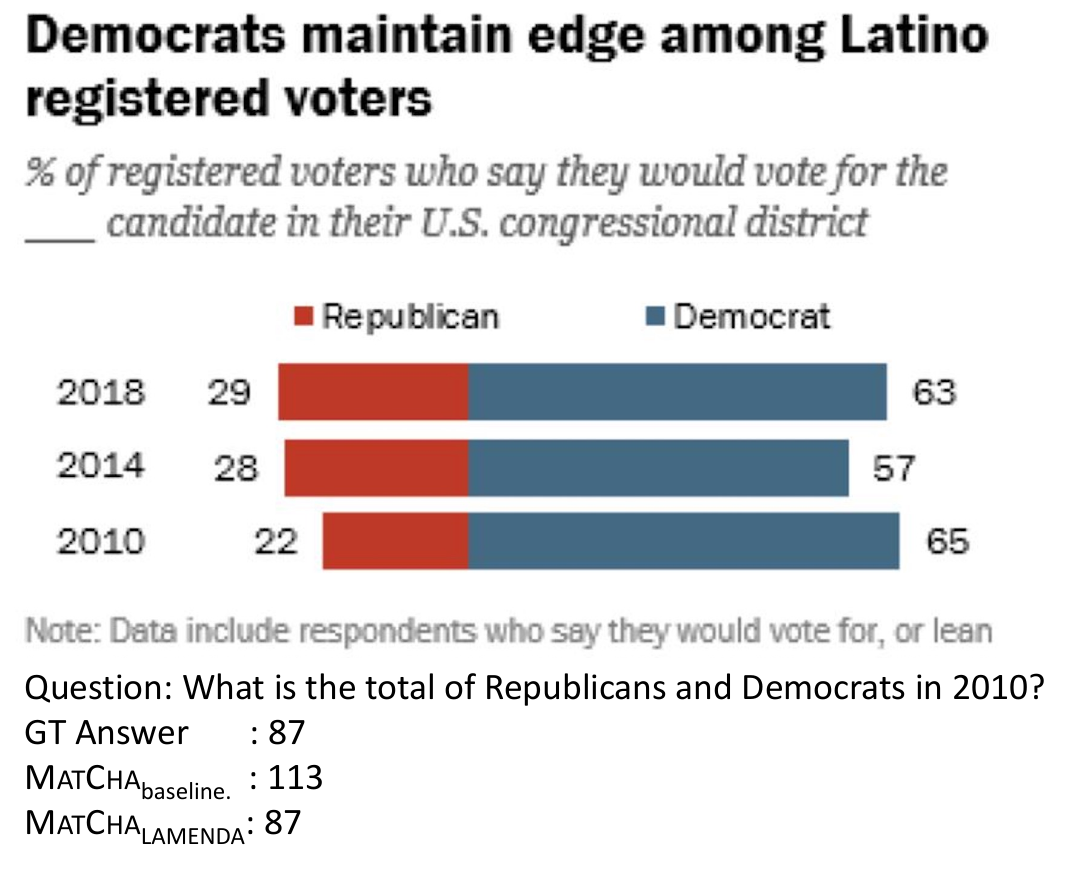}
  \caption{}
  \label{fig:sfig2}
\end{subfigure}
\caption{The baseline MatCha fails to correctly answer the question which needs visual understanding and then doing arithmetic operation, whereas MatCha fine-tuned with our data is able to correctly answer it.}
\label{fig:fig}
\end{figure}

% \begin{figure}[h]
% \begin{center}
%     % \fbox{\rule{0pt}{2in} \rule{.9\linewidth}{0pt}}
%     \includegraphics[width=0.75 \linewidth]{media/Analysis_1.pdf}
%     \caption{The baseline MatCha fails to correctly answer the question which needs visual understanding and then doing arithmetic operation, whereas MatCha fine-tuned with our data is able to correctly answer it.}
%     \label{fig:matcha_before_after}
% \end{center}
% \end{figure}

% \section{Examples of generated questions and answers}

\end{document}